\documentclass[letterpaper, 10 pt, conference]{ieeeconf}  

\IEEEoverridecommandlockouts                              

\usepackage{graphics} 
\usepackage{amsmath} 
\usepackage{amssymb}  
\usepackage{graphicx}
\usepackage{subfig}
\usepackage{algpseudocode}
\usepackage{algorithm}
\usepackage{url}
\usepackage[export]{adjustbox}
\usepackage{comment}
\usepackage{orcidlink}

\title{\LARGE \bf Simulation-based Planning of Motion Sequences for Automated Procedure Optimization in Multi-Robot Assembly Cells}

\begin{document}

\thispagestyle{empty} 
\null 
\vfill 
\begin{center} 
\begin{minipage}{0.8\textwidth} 
© 2026 IEEE.  Personal use of this material is permitted.  Permission from IEEE must be obtained for all other uses, in any current or future media, including reprinting/republishing this material for advertising or promotional purposes, creating new collective works, for resale or redistribution to servers or lists, or reuse of any copyrighted component of this work in other works.. 
\end{minipage} 
\end{center} 
\vfill 
\null 
\clearpage

\author{Loris~Schneider\orcidlink{0009-0009-1707-1124}, Marc~Ungen \orcidlink{0000-0002-9434-6007}, Elias~Huber \orcidlink{0009-0008-4190-140X}, Jan-Felix~Klein \orcidlink{0000-0002-3704-9567}%
\thanks{Loris Schneider, Elias Huber and Jan-Felix Klein are with the Institute for Material
Handling and Logistics (IFL), Karlsruhe Institute of Technology, 76131,
Germany. \texttt{loris.schneider@kit.edu}.}%
\thanks{Marc Ungen and Elias Huber are researchers at Bosch Corporate Research, Robert Bosch
GmbH, 71272 Renningen, Germany. \texttt{marc.ungen@de.bosch.com}}%
}
\maketitle
\thispagestyle{empty}
\pagestyle{empty}

\begin{abstract}
Reconfigurable multi-robot cells offer a promising approach to meet fluctuating assembly demands. However, the recurrent planning of their configurations introduces new challenges, particularly in generating optimized, coordinated multi-robot motion sequences that minimize the assembly duration.
This work presents a simulation-based method for generating such optimized sequences. The approach separates assembly steps into task-related core operations and connecting traverse operations. While core operations are constrained and predetermined, traverse operations offer significant potential for optimization. Therefore, the scheduling of core operations is formulated as an optimization problem, while feasible traverse operations are incorporated through a decomposition-based motion planning strategy.
Several solution techniques are explored, including a sampling heuristic, tree-based search and gradient-free optimization. For motion planning, a decomposition method is proposed that identifies specific subproblems in the schedule, which can be solved independently with modified centralized path planning algorithms.
The proposed method generates collision-free multi-robot assembly procedures that outperform a baseline relying on decentralized, robot-individual motion planning. Its effectiveness is demonstrated through simulation experiments.

\end{abstract}

\section{INTRODUCTION}

Fluctuations in resource availability and increasing demand for customized products accelerate the development of adaptive assembly systems. A key characteristic of such systems is their ability to adjust functionality within a fixed structural configuration \cite{el_maraghy_2009}. Multi-robot cells present a promising  realization of this paradigm, as general-purpose robotic manipulators enable task execution and parallelization, potentially reducing  overall assembly durations.

Realizing this potential requires optimizing both the structural and procedural configuration of the cell \cite{ungen_assembly}. The structural configuration specifies the available machine modules, manipulators and their arrangement in the workspace. The procedural configuration defines the execution logic that, together with the structural configuration, results in the capability to assemble a specific product with a specific performance. The procedural configuration consists of timed low-level control signals for the involved robots, tools, and machine modules. Optimizing such procedural configuration requires solving several nested problems, including task allocation, scheduling and motion planning.

\begin{figure}
    \centering
    \includegraphics[width=0.9\linewidth]{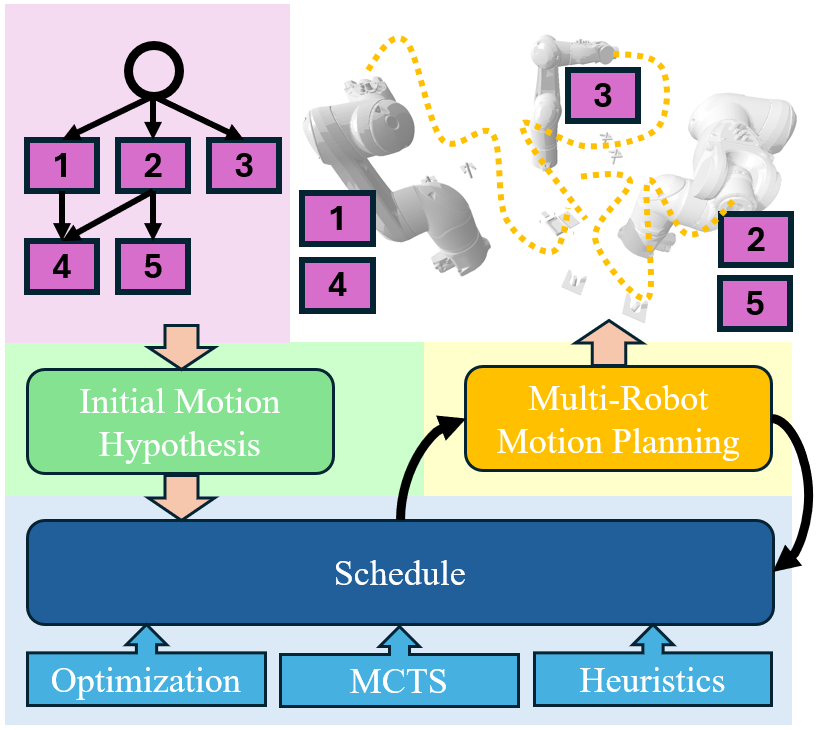}
    \caption{Overview of our approach. For an assembly task defined by an assembly precedence graph and an assignment of assembly steps to robots, initial motion hypotheses are generated. Those are arranged into an execution schedule which is optimized using three different optimization paradigms. Promising candidates are refined by a custom multi-robot motion planning method, allowing for complex multi-robot motions and achieving lower assembly durations.}
    \label{fig:teaser}
\end{figure}

To support systematic analysis, we use the following taxonomy to characterize procedural configurations of multi-robot assembly cells:\\
\textbf{Assembly task:} An assembly task represents the product-driven objective independent of any particular structural and procedural configuration. It is modeled as an assembly precedence graph specifying required assembly steps and their precedence constraints.\\
\textbf{Assembly step:} Assembly steps are the nodes of the precedence graph encapsulating specified operations that contribute towards the assembly task. Allocating these steps to specific robots yields a high-level execution schedule.\\
\textbf{Action:} Assembly steps are realized through sequences of low-level actions that depend on the structural and procedural configuration. These actions range from end-effector commands to different types of motion actions. Depending on the system design, both finer and coarser levels of action decomposition are possible.

This work addresses the problem of determining and scheduling efficient motion actions for a given structural configuration and a predefined assignment of assembly steps to robots. The goal is to generate collision-free motion instructions that ensure task execution while minimizing the total assembly duration.

Figure \ref{fig:teaser} provides an overview of our approach. Starting from an assembly precedence graph and an assignment of assembly steps to robots, we decompose the assembly steps into operations for which we create initial motion hypotheses. Those provide a lower bound for motion duration and an approximation of required spatiotemporal occupancy. From this, we create a high level schedule which we optimize using heuristics, gradient-free optimization methods or an adaptation of Monte Carlo Tree Search (MCTS) \cite{hutchison_bandit_2006}. Promising candidate schedules are refined and evaluated using a custom multi-robot motion planning scheme, which provides circular feedback to the schedule optimization. The result is an optimized schedule of complex multi-robot motion sequences, achieving faster assembly durations.

\section{Related Research}
\label{sec:related_work}
This section reviews relevant research on motion planning, multi‑modal motion planning, and task and motion planning.

\subsubsection{Motion planning}
Robotic motion planning addresses the generation of feasible, collision-free motions in the presence of obstacles. To account for differential constraints, it is often decomposed into path generation and subsequent time-parameterization of joint velocities and accelerations \cite{kant_toward_1986}. Pham and Pham propose a method for computing time-optimal joint velocity and acceleration trajectories \cite{pham_toppra_2018}.

If obstacle trajectories are known, time can be added to the configuration space $C$, yielding the space-time state space $X = C \times T$. Corresponding planning approaches are presented in \cite{sintov_time_based_2014, grothe_st_rrt_2022}. In multi-robot systems, the dimensionality of the joint configuration space increases planning complexity. Centralized methods plan in the combined $C$-space, often using prioritization or abstraction \cite{Erdmann_lozano_perez_1987, shkolnik_path_2009, orthey_multilevel_2020}, whereas decentralized methods plan sequentially for individual robots and trade completeness for computational efficiency \cite{van_den_berg_centralized_2009}.

\subsubsection{Multi-modal motion planning}
Multi-modal motion planning addresses problems in which the robot's kinematic structure changes during execution. Such problems can be represented by kinematic graphs $G = (V, E)$, where vertices denote rigid bodies and joints and edges encode attachments and relative transformations. Each state defines a mode with distinct kinematic and collision constraints. Planning across modes therefore requires separate configuration-space representations and connections at mode switches \cite{hauser_multi-modal_2010, hauser_random_mmmp_2011}.

\subsubsection{Task and motion planning}
Task and motion planning (TAMP) combines classical planning in a discrete state space $S$ with continuous or multi-modal motion planning. The resulting hybrid discrete-continuous problems are often formalized as hybrid constraint satisfaction problems (H-CSPs). Garrett et al. distinguish sequencing-first, satisfaction-first, and hybrid solution strategies \cite{garrett_integrated_2021}. Sequencing-first approaches first generate a task skeleton and then solve the associated motion-planning problems. Satisfaction-first approaches precompute feasible motions and mode transitions to constrain task planning. Hybrid approaches combine both principles.

\subsection{Related work}

Hartmann et al. present a multi-robot TAMP approach for long-horizon construction tasks using mobile manipulators \cite{hartmann_long-horizon_2023}. Assembly steps are allocated and executed iteratively with locally optimized asynchronous motions, yielding linear scalability with respect to the number of robots.

Chen et al. minimize assembly duration by constructing robot-specific annotated roadmaps that encode object grasps and connecting motions \cite{chen_cooperative_2022}. Possible inter-robot collisions are identified by comparing the individual roadmaps and are explicitly annotated. The global procedure is then synthesized from this roadmap information, allowing alternative grasps and motions to be considered during optimization. However, the collision annotation process is computationally expensive, which limits applicability in adaptive systems.

Ungen et al. propose a method for assigning assembly steps to robots and sequencing them to reduce the overall assembly duration in reconfigurable multi-robot assembly cells \cite{ungen_assembly}. Their approach relies on preplanned robot motions to efficiently optimize the temporal arrangement of assembly steps, but it does not consider coordinated multi-robot motion, leaving additional optimization potential unexploited.

Zhang et al. address multi-robot pick-and-place tasks with a sequencing-first TAMP approach \cite{zhang_multi-robot_2023}. Candidate plan skeletons are generated and partially instantiated, after which refinement is performed using MCTS. The search favors skeletons with fewer object relocations, thereby indirectly reducing execution time.

Mateu-Gomez et al. introduce a multi-robot TAMP approach that integrates motion planning directly into task-level optimization \cite{mateu-gomez_multi-arm_2024}. The workspace is discretized into a grid, and feasible, collision-annotated joint configurations are defined for each grid point, enabling a formulation as a Markov Decision Process. However, the optimization is restricted to the predefined joint configurations associated with the discretization.

\begin{figure*}[!t]
\centering
\subfloat{\includegraphics[angle=90,width=\textwidth]{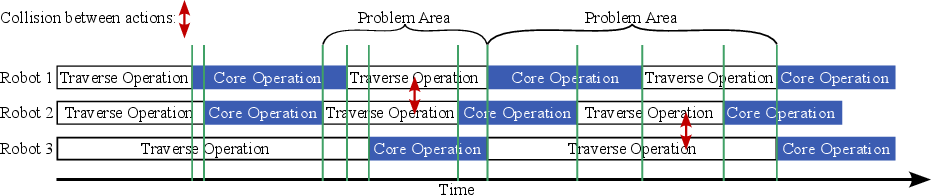}}
\caption{Identification of problem areas in a schedule based on collisions of traverse operations.}
\label{problem_areas}
\end{figure*}

\section{Methodology}
\label{sec:methodology}
Building on the work of \cite{ungen_assembly}, we assume a well-defined assembly precedence graph, a multi-robot cell with a fixed structural configuration, and a predefined assignment of assembly steps to robots. The following assumptions hold:
\begin{enumerate}
    \item All required components and tools are available with known initial and goal locations.
    \item Each assembly step specifies the components and tools it requires.
    \item Assigned robots can reach all relevant locations, no handovers needed.
    \item Each robot has a collision-free escape configuration.
\end{enumerate}

\subsection{Problem decomposition}
Each assembly step decomposes into a sequence of low-level actions, which are grouped into core operations and traverse operations, see Table \ref{tab:action_group_classes}.
\begin{itemize}
    \item \textbf{Core operations} represent the productive parts of an assembly step.
    \item \textbf{Traverse operations} consist of a single motion action that moves the robot to the start configuration of the next core operation.
\end{itemize}
Thus, each assembly step can be viewed as an alternating sequence of traverse and core operations.

\begin{table}[!t]
\caption{Assignment of actions to traverse and core operations.}
\label{tab:action_group_classes}
\centering
\small
\setlength{\tabcolsep}{4pt}
\renewcommand{\arraystretch}{1.05}
\begin{tabular}{l|l}
\textbf{Operation} & \textbf{Actions of assembly step} \\
\hline
Traverse operation & move to object \\
Core operation & open gripper, approach, close gripper \\
Traverse operation & move to target \\
Core operation & open gripper, depart
\end{tabular}
\end{table}

The first step of the approach determines suitable motions for all core operations. These operations typically involve precise manipulations subject to kinematic or force constraints. They are therefore preplanned as time-parametrized joint trajectories with fixed durations. In this work, constraints on core motions are expressed as end-effector waypoints. Corresponding joint trajectories are generated using inverse kinematics followed by a time-optimal path parametrization \cite{pham_toppra_2018}, ensuring smooth execution with zero velocity at start and end.
Once defined, core operations are considered fixed motion sequences. Accordingly, the remaining problem is divided into two coupled subproblems:

\begin{enumerate}
    \item \textbf{Scheduling}: determine the start times of the core operations.
    \item \textbf{Motion Planning}: generate collision-free traverse motions connecting successive core operations.
\end{enumerate}

The proposed method adopts a sequencing-first strategy. First, a schedule of core operations is generated, as shown in Figure \ref{problem_areas}, with each core operation represented as a fixed-duration block. The white intervals between these blocks correspond to traverse operations, within which feasible collision-free motions are planned. These motions are subsequently fed back into the scheduling process, enabling iterative refinement of the overall solution. In the following, the scheduling method is presented first, followed by the motion planning approach.

\subsection{Scheduling}
\label{sec:scheduling}
A schedule for the $N$ core operations $ O^{\mathrm{core}}= [o_1^{\mathrm{core}}, o_2^{\mathrm{core}}, ..., o_N^{\mathrm{core}}]$ is represented by a $N$-dimensional vector $\theta$, where each entry denotes the start time of a core operation. The scheduling problem is formulated as: 
\begin{align}
    \min_\theta \quad &d(\theta) \label{scheduling_problem_base}\\
    \textrm{s.t.} \quad & c(\theta) = 0 \label{core_collision_constraint}\\
    & p(\theta) \geq 0 \label{precedence_constraints}\\ 
    & m(\theta) \geq 0 \label{motion_planning_constraints}
\end{align}
Here, $d(\theta)$ is the resulting assembly duration. The constraint in equation \eqref{core_collision_constraint}  requires zero collisions during core operations, while the constraint in equation \eqref{precedence_constraints} enforces precedence relations between core operations. Finally, the constraint in equation \eqref{motion_planning_constraints} requires sufficient time between consecutive core operations on the same robot to allow collision-free traverse motions.

Evaluating \eqref{motion_planning_constraints} requires solving the full motion planning problem as described in Section \ref{sec:motion_planning}, which is computationally demanding. To reduce complexity during initial scheduling, this constraint is approximated by estimating a lower bound on the duration of each traverse operation.
This estimate is obtained by generating initial motion hypotheses by computing a path for each robot individually, shortening and applying time-optimal parametrization \cite{pham_toppra_2018}. We express this relaxed constraint as:
\begin{align}
    m'(\theta) \geq 0 \label{motion_planning_constraints_relaxed}
\end{align}
Together with the known durations of core operations, these estimates allow the computation of a lower bound $d_{\textrm{min}}$ on the total assembly duration. This bound respects precedence constraints but does not consider possible collisions during traverse operations.

\subsubsection{Sampling heuristic}
\label{sampling_heuristic}

To obtain a feasible schedule $\theta$, satisfying constraints (\ref{core_collision_constraint}), (\ref{precedence_constraints}), and (\ref{motion_planning_constraints_relaxed}) while keeping the assembly duration short, two additional relaxations are introduced. First, time is discretized into fixed-length intervals. Second, a maximum assembly duration $d_{\textrm{max}}$ is defined, either as a multiple of the lower bound $d_{\textrm{min}}$ or as a tunable parameter.

Based on these relaxations, a sampling heuristic is used to construct candidate schedules, as summarized in Algorithm \ref{alg:sampling_heuristic}. In each iteration, the algorithm determines the set of \textit{primed} core operations, i.e., core operations whose predecessors have all been scheduled or that have no predecessors. Then a primed core operation is selected at random. For this core operation, a feasible start interval is computed, bounded by the earliest start time, which is determined by the latest predecessor completion, and the latest start time that still permits all required successor core and traverse operations to be scheduled within $d_{\textrm{max}}$.

\begin{algorithm}
    \caption{Scheduling by sampling heuristic}\label{alg:sampling_heuristic}
    \begin{algorithmic}[1]
        \State $O^{\mathrm{core}}_{\textrm{closed}} \gets \emptyset, O^{\mathrm{core}}_{\textrm{open}} \gets O^{\mathrm{core}}$
        \While{not $O^{\mathrm{core}}_{\mathrm{open}} = \emptyset$}
            \State $o^{\mathrm{core}} \gets getPrimedCoreOperation(O^{\mathrm{core}}_{\textrm{open}}, O^{\mathrm{core}}_{\textrm{closed}})$
            \State $I \gets getSchedulingInterval(o^{\mathrm{core}}, O^{\mathrm{core}}_{\textrm{open}}, O^{\mathrm{core}}_{\textrm{closed}})$
            \If{$I = \emptyset$}
                \State start over
            \EndIf
            \State $P \gets getSelectionProbabilities(I)$
            \State $assignStartTimeStep(o^{\mathrm{core}}, I, P)$
            \While{$collision(o^{\mathrm{core}}, O^{\mathrm{core}}_{\mathrm{closed}})$ is true }
                \State $I.pop(s)$
                \If{$I = \emptyset$}
                    \State start over
                \EndIf
                \State $P \gets getSelectionProbabilities(I)$
                \State $assignStartTimeStep(o^{\mathrm{core}}, I, P)$
            \EndWhile
            \State $O^{\mathrm{core}}_{\textrm{closed}}.push(o^{\mathrm{core}}), O^{\mathrm{core}}_{\textrm{open}}.pop(o^{\mathrm{core}})$
        \EndWhile
    \end{algorithmic}
\end{algorithm}

A probability distribution biased toward earlier start times is defined over the feasible interval to promote short assembly durations. A start time is sampled from this probability distribution, assigned to the selected core operation, and checked for collisions with already scheduled core operations. If a collision is detected, the sampled time step is removed from the scheduling interval, the probability distribution is updated to fit the now reduced scheduling interval, and the process is repeated. If the scheduling interval becomes empty, the algorithm restarts, which was found to be more efficient than backtracking. The heuristic therefore generates schedules that satisfy constraints (\ref{core_collision_constraint}), (\ref{precedence_constraints}), and (\ref{motion_planning_constraints_relaxed}).

\subsubsection{Gradient-free optimization}
\label{gradient_free_optimization}

As a second scheduling approach, the problem is reformulated to permit the use of gradient-free optimization methods. The collision constraint \eqref{core_collision_constraint} is transferred into the objective as a penalty term, such that only the precedence constraint \eqref{precedence_constraints} and the relaxed motion-feasibility constraint \eqref{motion_planning_constraints_relaxed} must be satisfied explicitly. The resulting objective is

\begin{align}
    \label{objective_function_gradient_free}
    \min_\theta \quad &d(\theta) + \mu_1 c^2(\theta) + \mu_2(\max\{0, d(\theta) - d_{\textrm{max}}\})^2
\end{align}

Here, $c(\theta)$ denotes the number of collisions in the schedule. The third term penalizes schedule extensions beyond $d_{\textrm{max}}$, preventing collision avoidance through excessively long schedules. The weights $\mu_1$ and $\mu_2$ determine the relative influence of collision and duration penalties and are typically chosen such that $\mu_2 \gg \mu_1 \gg 1$.

The considered gradient-free methods are Simulated Annealing (SA) \cite{kirkpatrick_optimization_1983} with the statistical cooling strategy proposed in \cite{aarts_statistical_1985,doring_multi-robot_2005}, Particle Swarm Optimization (PSO) \cite{eberhart_kennedy_part_swarm}, and an Evolutionary Algorithm (EA). All methods are adapted to satisfy constraints \eqref{precedence_constraints} and \eqref{motion_planning_constraints_relaxed}. Initialization is performed either from randomly generated schedules or from the sampling heuristic in Section \ref{sampling_heuristic}, which provides feasible, penalty-free starting points.

Motion-planning feedback is incorporated into the objective as described in Section \ref{feeding_back_motion_planning_results}. The parameter settings are given in Section \ref{sec:evaluation}.

\subsubsection{Decision tree}
\label{formulation_as_decision_tree}
\begin{figure}[!t]
\centering
\subfloat[]{\includegraphics[width=0.08\textwidth, valign=t]{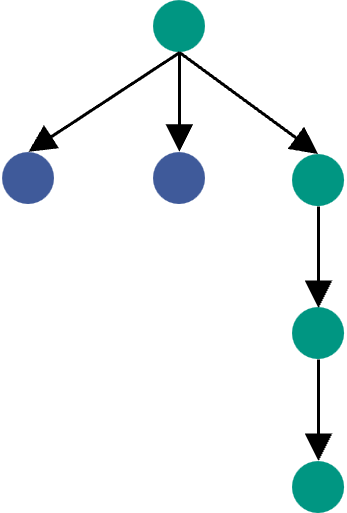}\label{mcts_g_a}}
\qquad
\subfloat[]{\includegraphics[width=0.08\textwidth, valign=t]{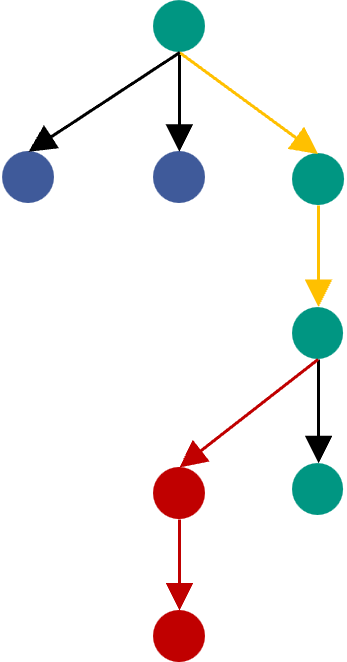}\label{mcts_g_b}}
\qquad
\subfloat[]{\includegraphics[width=0.08\textwidth, valign=t]{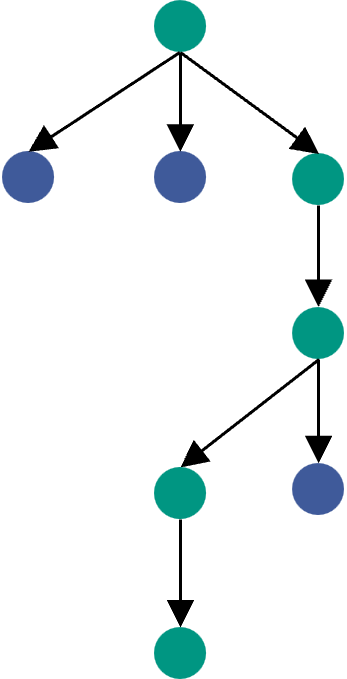}\label{mcts_g_c}}
\caption{Exploration of the decision tree by MCTS-G. (a) The current best path is marked by green nodes. (b) According to UCT \cite{hutchison_bandit_2006}, path (yellow) is followed until a random leaf node is reached. From there, a playout (red) starts until a terminal node is reached. (c) The entire playout path is added to the explored tree.}
\label{mcts_g}
\end{figure}
The scheduling problem can also be formulated as a decision tree.
Each core operation corresponds to a node, arranged in a predetermined scheduling order that respects the precedence constraints.
The possible time steps within the operation’s scheduling interval are represented as outgoing edges. A complete schedule is therefore a path from the tree root to a leaf node, while terminal nodes may correspond to valid schedules or invalid ones resulting from collisions or infeasible timing. This representation implicitly encodes all feasible and infeasible schedules. Searching the decision tree enables systematic exploration of the scheduling space.
A widely used method for exploring decision trees is Monte Carlo Tree Search (MCTS). MCTS iteratively applies four main phases: selection, expansion, simulation and backpropagation.

During the selection phase, child nodes are commonly chosen by maximizing the Upper Confidence Bounds Applied to Trees (UCT) score \cite{hutchison_bandit_2006}, balancing the exploration-exploitation trade-off.
Because standard MCTS assumes maximization of a scalar reward in $[0,1]$, we design a reward function by adapting and normalizing the  objective function:
\begin{align}
     r(\theta) = \begin{cases}
        \frac{\lambda_1 N_{\mathrm{free}}}{d_{\textrm{max}} - d_{\textrm{min}} + \lambda_1 N}, & \mathrm{if} \quad c(\theta) \neq 0\\
        0, & \mathrm{if} \quad d(\theta) > d_{\textrm{max}}\\
        \frac{d_{\textrm{max}} - d(\theta) + \lambda_1 N}{d_{\textrm{max}} - d_{\textrm{min}} + \lambda_1 N}, & \mathrm{else}\\
    \end{cases} \label{result_value_function}
\end{align}
\begin{figure*}[t]
\centering
\subfloat{\includegraphics[angle=90,width=0.7\textwidth]{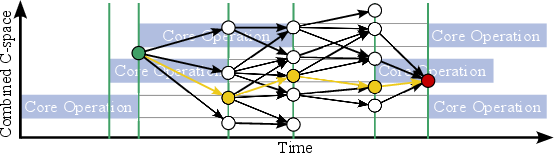}}
\caption{A cross-sectional roadmap with a guide path.}
\label{roadmap}
\end{figure*}
Here, $N$ is the number of all core operations and $N_{\mathrm{free}}$ is the number of successfully scheduled core operations without collisions.
The parameter $\lambda_1$ balances the importance of collision-free scheduling and total assembly duration.

During simulation, MCTS may perform \textit{light} playouts (random decisions) or \textit{heavy} playouts using a more informed strategy. The sampling heuristic from Section \ref{sampling_heuristic} provides such a heavy-playout strategy, with core operations processed according to the predetermined scheduling order.

Because classical MCTS is primarily designed for adversarial game settings, a modified variant MCTS-Greedy (MCTS-G) is used. This variant stores the best observed result in each node rather than the average. After playouts, we add the entire playout path to the explored tree instead of a single child. Selection is adapted to possibly continue past leaf nodes with at least one explored child, stopping only with a defined probability.
These adaptations bias the search toward promising regions of the decision tree while still permitting exploration.  
Figure \ref{mcts_g} illustrates how the explored subtree grows, how promising paths are pursued, and how newly discovered paths that improve upon the current best solution become the new reference paths.

\subsection{Motion Planning}
\label{sec:motion_planning}
All scheduling methods in Section \ref{sec:scheduling} enforce only the relaxed motion constraint \eqref{motion_planning_constraints_relaxed}. To evaluate the stricter constraint \eqref{motion_planning_constraints}, a dedicated multi-robot motion-planning method is introduced.

\subsubsection{Identification of Problem Areas}
\label{identifiaction_of_problem_areas}
The method starts by identifying \textit{problem areas}, see Figure \ref{problem_areas}. 
The schedule is initially segmented by identifying all start and end times of core operations (green vertical lines).
Since mode switches occur only during core operations, each resulting section contains traverse operations executed under fixed modes. 

Sections consisting solely of core operations require no further planning. Among the remaining sections, we identify problem areas by clustering all sections containing traverse operations whose initial motion hypotheses result in collisions. Sections not included in a problem area can be executed using their initial motion hypotheses.
The exemplary schedule in Figure \ref{problem_areas} shows two problem areas resulting from collisions between traverse operations.

\subsubsection{Creating Cross-Sectional Roadmaps}
\label{finding_guide_paths}
Each problem area is resolved by constructing a \textit{cross-sectional roadmap}, as shown in Figure \ref{roadmap}. Such a roadmap constitutes a graph in the combined $C$-space of all robots that allows the generation of solution skeletons. 

At each section boundary, collision-free configurations are sampled from the combined $C$-space of robots performing traverse operations at this time step. Robots executing core operation are treated as static obstacles with known joint configurations. 
To ensure that only reachable configurations are sampled, we calculate the reachable joint angle intervals for each individual robot, constraining the sampling space to feasible regions. 
Each collision‑free configuration at a section boundary constitutes a node in the roadmap and is subsequently connected by an edge to configurations at the next boundary whenever the transition is executable under velocity and acceleration limits and available time. Additional samples are generated until at least one valid connection is found or an iteration limit is reached. The resulting directed graph, as illustrated in Figure \ref{roadmap}, constitutes the cross‑sectional roadmap. We apply edge weights corresponding to distance in $C$-space.

A \textit{guide path}, highlighted in yellow in Figure \ref{roadmap} is obtained using Dijkstra's algorithm \cite{dijkstra_note_1959}. If no such path exists, the roadmap is incrementally expanded with additional samples until a guide path becomes available.

\subsubsection{Motion Planning along guide paths}
\label{motion_planning_alogn_guide_paths}
Each edge in the guide path constitutes an individual motion-planning problem between two configurations.
Robots involved in core operations act as dynamic obstacles with known states over time. Each subproblem is solved in the combined $X$-space of the traversing robots using PRM \cite{kavraki_probabilistic_1996} and RRT-Connect \cite{rrt_connect}, which we constrain to respect velocity and acceleration limits. If a subproblem cannot be solved, the corresponding edge is removed from the roadmap and a new guide path is computed. If no guide path exists, the roadmap is augmented with additional samples. Solved subproblems are concatenated to form a valid trajectory across the entire problem area. 

\subsubsection{Feeding back motion-planning results}
\label{feeding_back_motion_planning_results}
Some problem areas may not admit a solvable guide path. To prevent indefinite searching, we limit the number of sampled configurations on section borders during the roadmap construction. Additionally, we impose a time limit for solving motion-planning subproblems. This results in two failure modes. Either roadmap construction failure by reaching the sampling limit or guide-path solving failure by reaching the time limit.
For both failure cases, the \textit{magnitude of failure} is quantified by counting the number of traverse operations ending before the critical time point. These counts are denoted as $A_R$ (roadmap failure) and $A_P$ (planning failure).

\begin{figure*}[!t]
\centering
\subfloat[]{\includegraphics[width=0.3\textwidth]{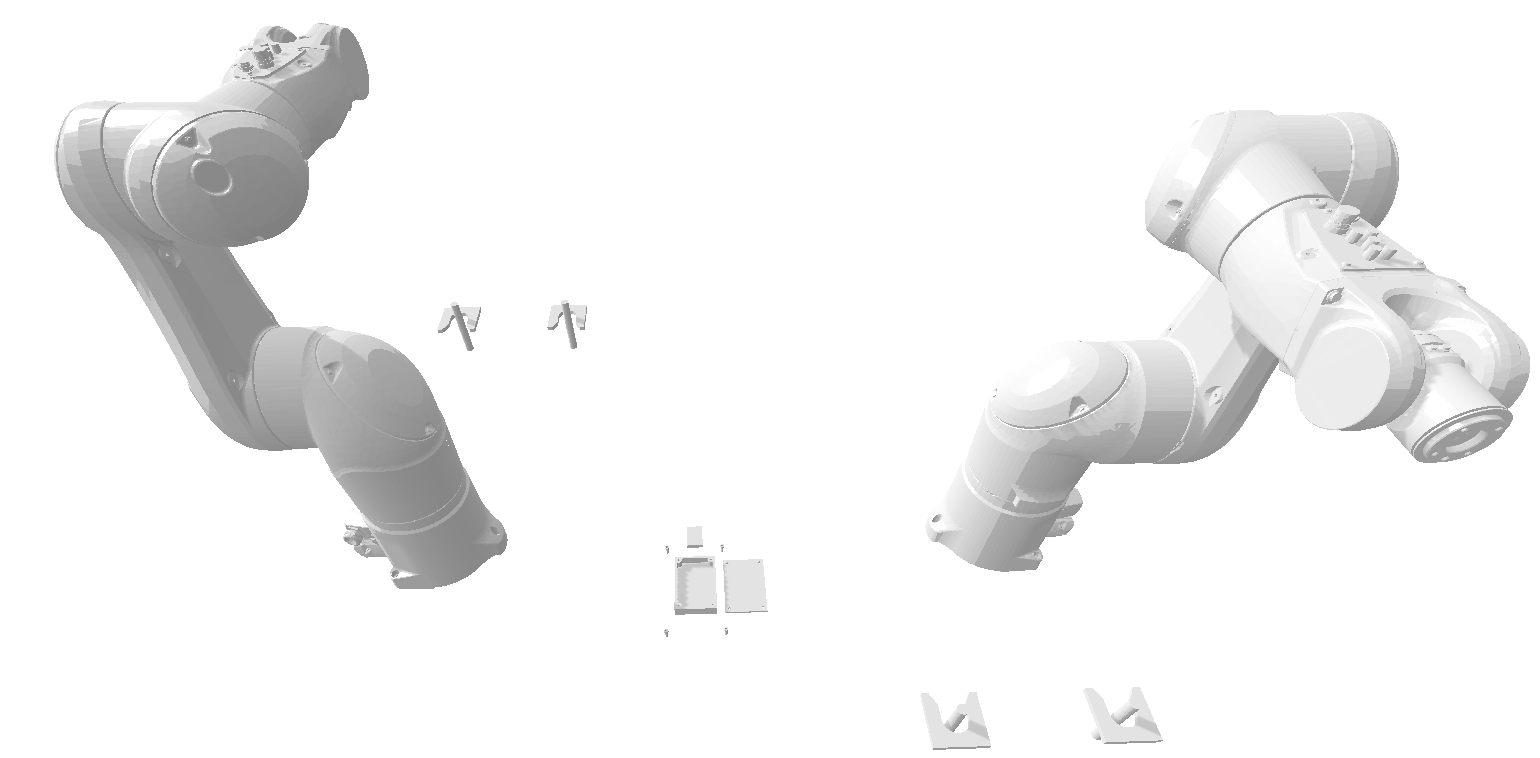}}
\hfill
\centering
\subfloat[]{\includegraphics[width=0.3\textwidth]{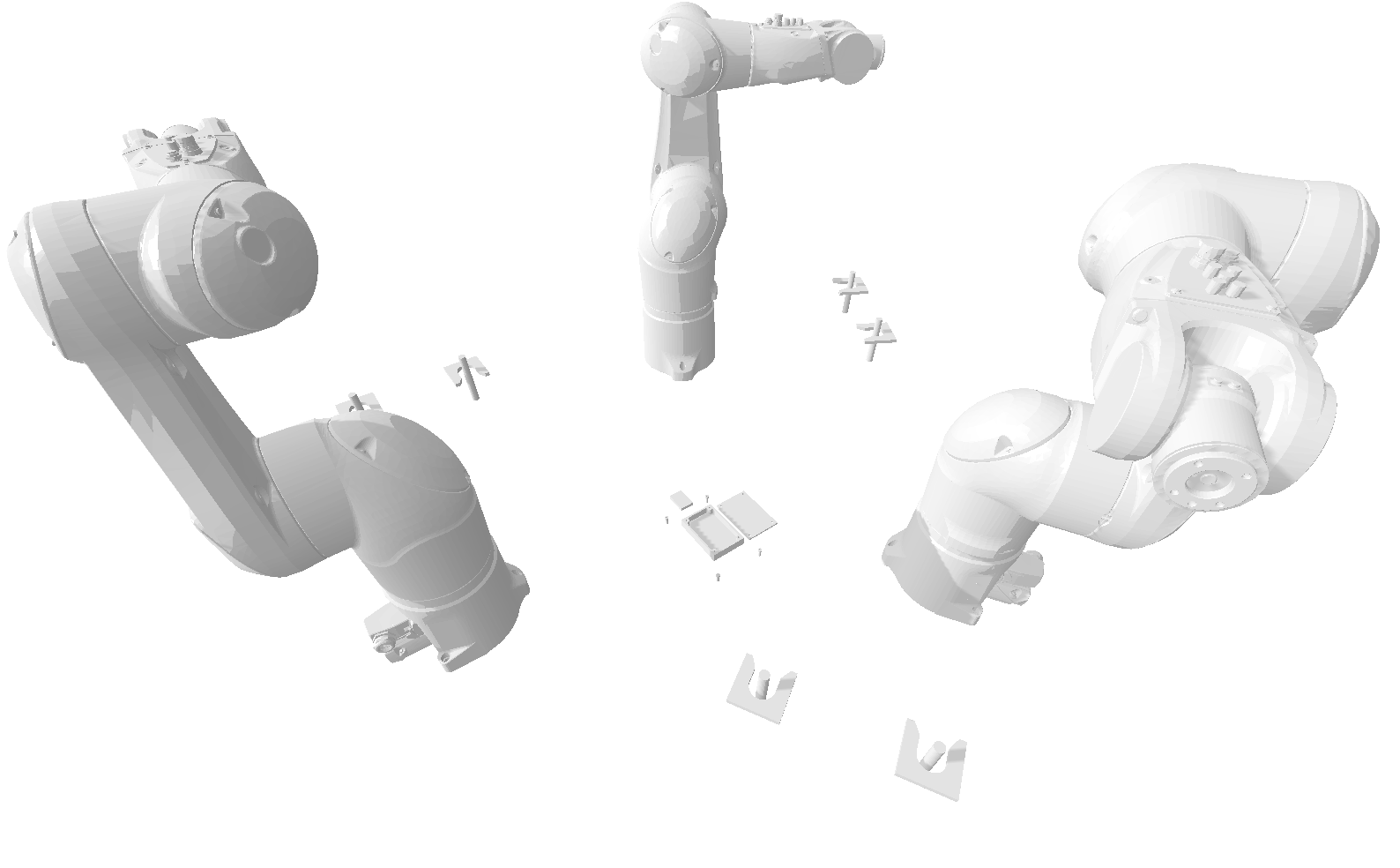}}
\hfill
\centering
\subfloat[]{\includegraphics[width=0.3\textwidth]{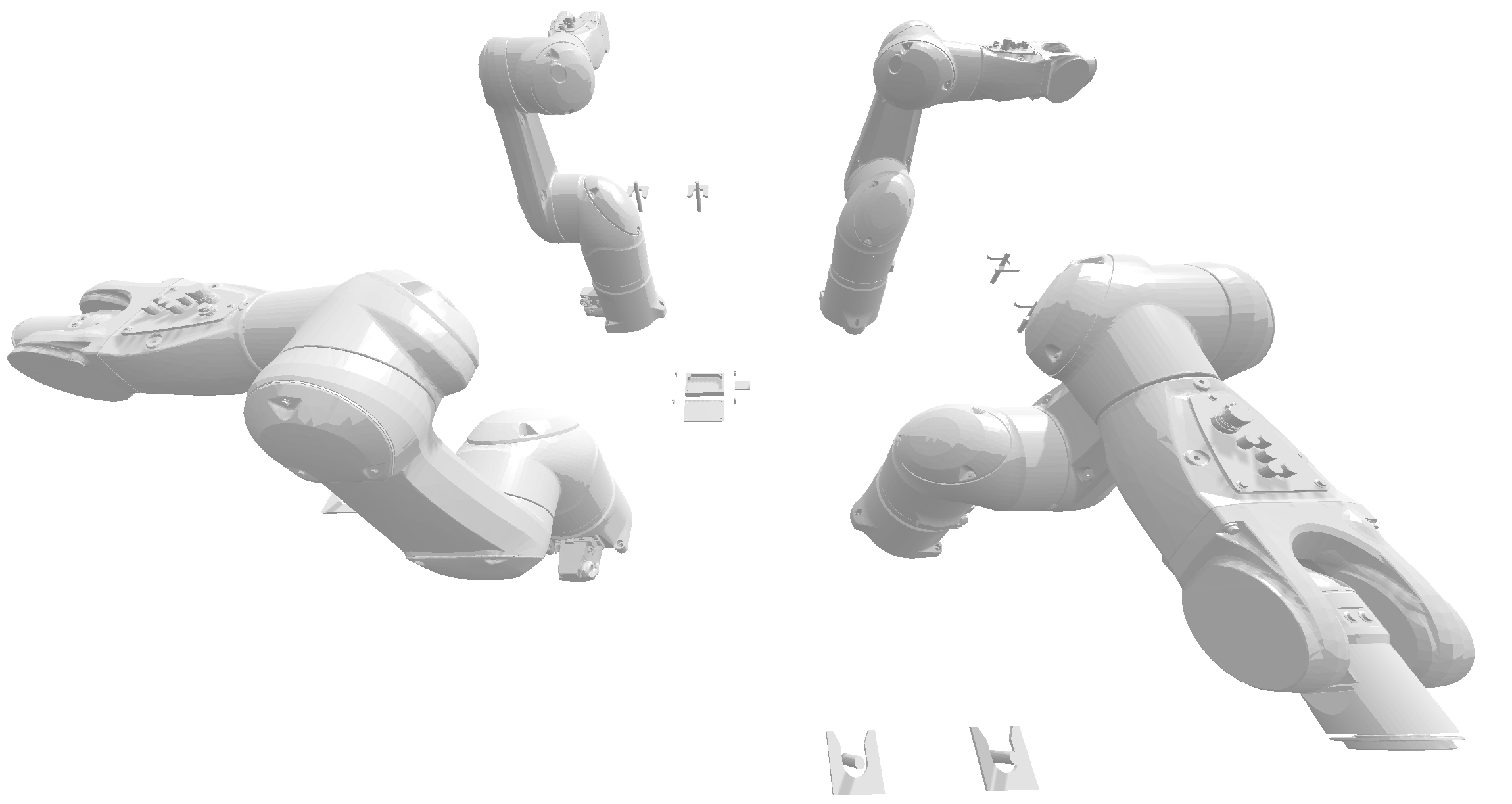}}
\hfill
\caption{Structural configurations with two (a), three (b) and four (c) robots.}
\label{task_2_settings}
\end{figure*}

These quantities influence the gradient-free optimization by extending the objective function (\ref{objective_function_gradient_free}) with weighted penalization terms
\begin{align*}
    - \mu_3A_P - \mu_4A_R \nonumber.
\end{align*}

Similarly, the result value function (\ref{result_value_function}) of the MCTS-based methods is extended:
\begin{align*}
    \frac{\lambda_1 N_{\mathrm{free}}}{d_{\textrm{max}} - d_{\textrm{min}} + \lambda_1 N + \lambda_2N + \lambda_3N} \nonumber
\end{align*}
if $c(\theta) \neq 0$ and
\begin{align*}
    \frac{d_{\textrm{max}} - d(\theta) + \lambda_1 N + \lambda_2A_P + \lambda_3A_R}{d_{\textrm{max}} - d_{\textrm{min}} + \lambda_1 N + \lambda_2N + \lambda_3N} \nonumber
\end{align*}
in the default case with the weights $\lambda_2$ and $\lambda_3$.\\

Finally, successfully solved guide paths undergo post-process using short-cutting, corner-cutting \cite{chen_sandros_1998}, partial short-cutting \cite{geraerts_creating_2007} and moving-average smoothing to improve trajectory quality.

\section{Evaluation}
\label{sec:evaluation}

An assembly task involving seven components to be assembled to a small box was designed to evaluate the presented scheduling and motion‑planning methods.
Precedence constraints ensure that placement operations precede lid fixation and allow screw insertions to occur in parallel. Experiments were conducted using structural configurations comprising two, three, and four robots, each with its own tool holder and tool set, see Figure \ref{task_2_settings}.

All experiments were performed in simulation, using PyBullet \cite{pybullet} for visualization, rigid body dynamics and collision checking. Adapted versions of PRM and RRT-Connect were used for path planning through the OMPL library \cite{ompl} via its Python API. All experiment runs were executed within virtual machines configured with three logical CPUs and 20GB RAM on a host system with two 10-core Intel Xeon E5-2690 v2 CPUs at 3.0 GHz and 255 GB RAM.

\subsection{Investigated methods}
\label{investigated_methods}
Each scheduling method was allocated 24 hours per experiment, and five independent runs were performed for each method and configuration.\\
The gradient-free optimization methods described in Section \ref{gradient_free_optimization} (SA, PSO and EA) were evaluated both with random initialization ("Rand"), and with an initialization via the sampling heuristic ("Init"). MCTS and MCTS-G were evaluated using the sampling heuristic for heavy playouts, random light playouts, and mixed playouts. For mixed configurations, the percentage of light playouts is indicated (e.g., "MCTS-G 10\%" corresponds to 10\% light playouts and 90\% heavy playouts).
All methods are compared against a baseline consisting of continuously sampling random schedules.

\begin{figure*}[!t]
    \centering
    \subfloat{\includegraphics[width=0.95\textwidth]{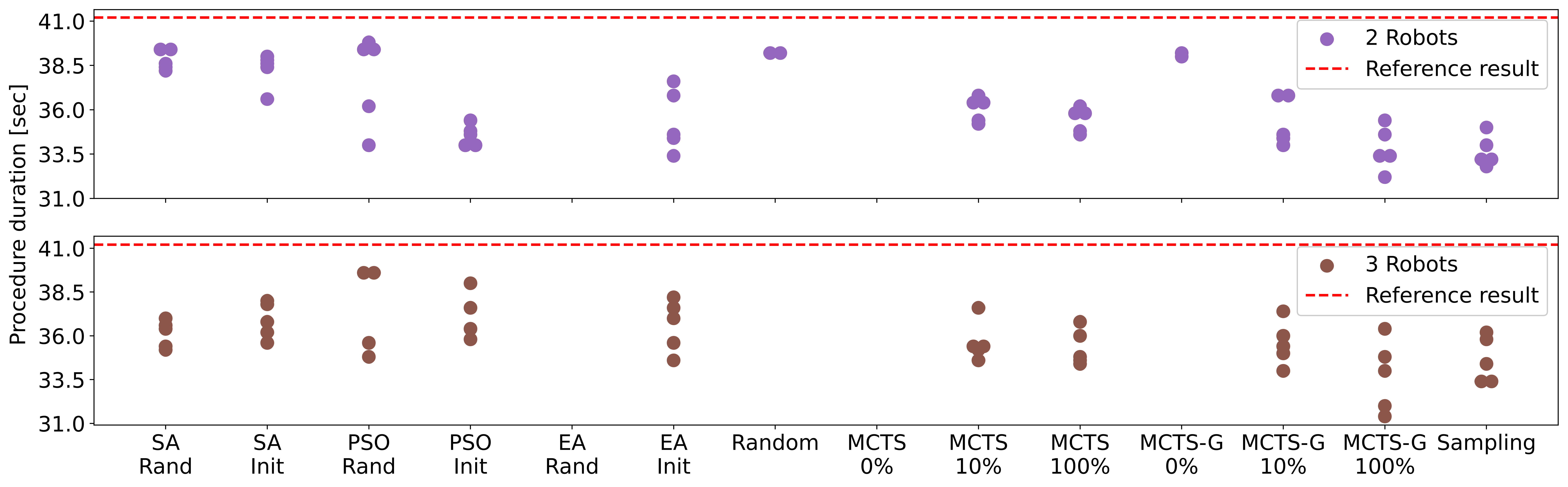}}
    \hfil
    \caption{Optimized solutions after five experiments.}
    \label{best_results_all}
\end{figure*}

\subsection{Parameters}
\label{parameters}
The maximum assembly duration was set to $d_{\mathrm{max}} = 40 s$ and the time discretization for scheduling was  $0.2 s$ seconds, yielding 200 discrete time steps. 

\subsubsection{Sampling heuristic parameters}
The main parameter of the sampling heuristic is the probability distribution used to select start times. 
For a scheduling interval $I=[t_0, ..., t_M]$, the probability of selecting time step $t_i$ is
\begin{align}
    p(t_i) = \frac{f(t_i - t_0)}{\sum_{j=0}^M f(t_j - t_0)}, \nonumber
\end{align}
where $f$ is a function over time $t$ that can be chosen as a hyperparameter. In the following, a  quadratically declining function is used: 
\begin{align}
    f(t) &= \frac{1}{(t+2)^2} \nonumber\\
\end{align}

\subsubsection{Gradient-free optimization parameters}
\paragraph{Simulated Annealing}
The statistical cooling strategy defined in \cite{aarts_statistical_1985} was used with $x_0 = 0.95$, $\delta = 0.2$, and $ m_0 = L = 20$.

\paragraph{Particle Swarm Optimization}
The velocity update suggested by \cite{shi_eberhart_pso} was used:
\begin{align}
    v^p_{i+1} = w^pv^p_i + c_{\mathrm{soc}}\epsilon_1(x^g_{\mathrm{best}} - x^p_i) + c_{\mathrm{cog}}\epsilon_2(x^{p}_{\mathrm{best}} - x^p_i)\nonumber
\end{align}
with parameter values $w^p = 0.9$, $c_{\mathrm{soc}} = 2$, $c_{\mathrm{cog}} = 1$, $N_p = 15$, $v_{\mathrm{min}} = -20$, and $v_{\mathrm{max}} = 20$.
The variable $N_p$ is the number of particles and $v_{\mathrm{min}}$ and $v_{\mathrm{max}}$ are bounds on the randomly chosen initial velocity.
\paragraph{Evolutionary Algorithm}
The EA used a population of $N_{\mathrm{EA}} = 15$ individuals with 5-point crossover. The seven best individuals produced 14 descendants per iteration. Mutation modified 20\% of values using Gaussian noise with $\sigma = 3$. The best individual of each iteration was preserved.\\
The extended objective function (\ref{objective_function_gradient_free}) used the following weights: $\mu_1 = 30$, $\mu_2 = 900$, $\mu_3 = 10$, and $\mu_4 = 1$.

\subsubsection{MCTS‑based method parameters}
For both MCTS variants, the result-value function used $\lambda_1 = 0.01$, $\lambda_2 = 1$, and $\lambda_3 = 0.1$.
The UCT exploration constant was set to $c = \sqrt{2}$.
For MCTS-G, the probability of terminating selection at a leaf node with explored children was set to 10\%.

\subsubsection{Motion-planning parameters}
The sampling attempt limit at each section boundary was set to 
$L_0 = 10$. The time limit for each motion-planning subproblem was set to $T_{max} = 30 \times n_r$ seconds, and the time limit for advancing the guide path was set to 15 minutes.

\subsection{Results}
Figure \ref{best_results_all} summarizes the optimized solutions obtained across all investigated methods for two and three robots. For the sampling heuristic, the best solution is shown for each run. "Random" denotes schedules generated uniformly at random, whereas  "Sampling" denotes the sampling heuristic with a quadratically declining probability distribution.

\begin{figure}[t]
    \centering
    \includegraphics[width=0.6\columnwidth]{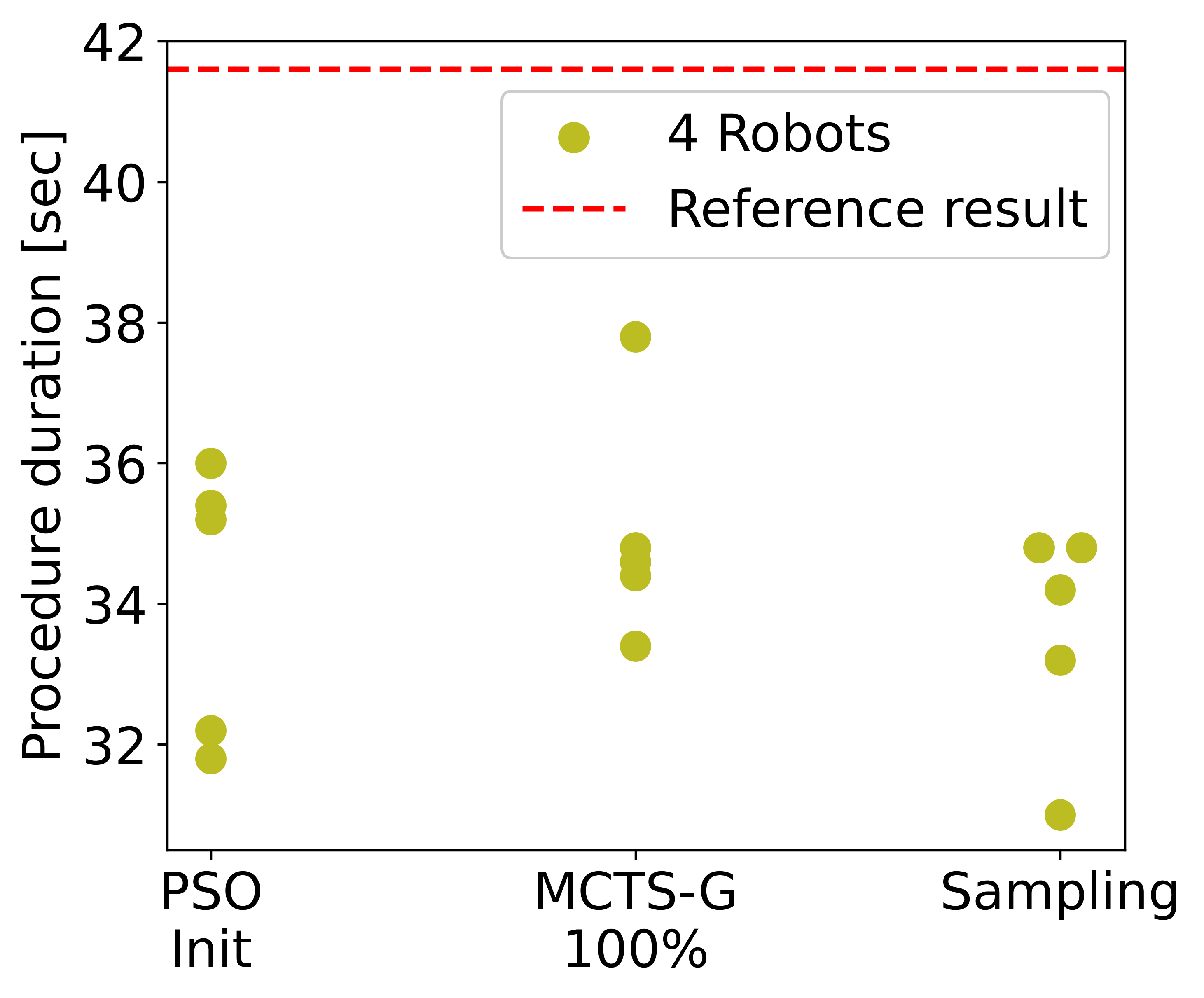} 
    \caption{Optimized solutions in a structural configuration with four robots after five experiments.}
    \label{best_results_b4p1}
\end{figure}

MCTS and MCTS-G with purely light playouts, and EA with random initialization failed to generate solutions in all runs for both two-robot and three-robot configurations. Across all experiments, no method emerged as uniformly superior.
To evaluate scalability to larger systems, three methods were applied to the four-robot configuration: PSO with initialization using the sampling heuristic, MCTS-G with purely heavy playouts, and the sampling heuristic. As shown in Figure \ref{best_results_b4p1}, all three methods reliably produced valid solutions. This confirms the applicability of the generated approach to structural configurations with four robots.

Compared with \cite{ungen_assembly}, which achieves assembly durations of 41.2s for the two- and three-robot configurations and 41.6s for the four-robot configuration, the proposed approach reduces assembly duration by approximately 10-20\% across all tested configurations. The corresponding reference values are indicated by the red dashed lines in Figures \ref{best_results_all} and \ref{best_results_b4p1}.


Computation times for successful motion planning are shown in Figure \ref{computation_time_total}. The boxplots summarize the distributions of runtime for all successfully solved motion-planning instances; the numbers above the boxplots indicate the corresponding sample counts. Median computation time and variability increase markedly with the number of robots, indicating non-linear scaling due to the growth of the combined configuration space.

\begin{figure}[t]
    \centering
    \includegraphics[width=0.6\columnwidth]{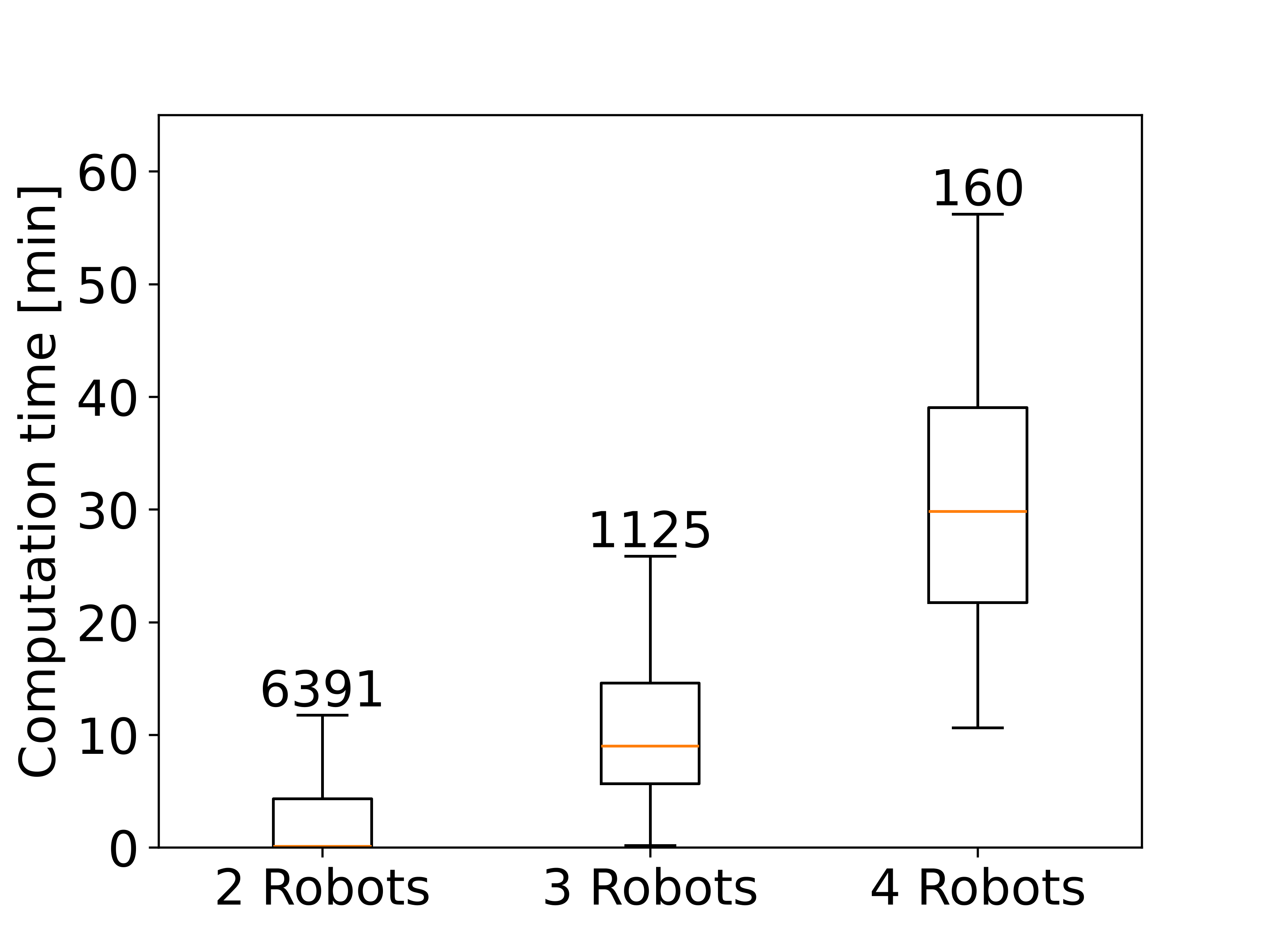}\label{computation_time_total}
    \caption{Computation time for successful motion planning.}
    \label{computation_time_total}
\end{figure}

\section{Conclusion and Limitations}
\label{sec:conclusion}
This paper presented an approach for minimizing assembly duration in reconfigurable multi-robot assembly cells under a fixed structural configuration and a fixed assignment of assembly steps. The method decomposes assembly steps into preplanned core operations and traverse operations connecting them. The scheduling of core operations is optimized using a sampling-based heuristic, a decision-tree formulation, and several gradient-free optimization methods.

To account for multi-robot interactions, a semi-centralized motion planning method was introduced. It identifies regions in which decentralized planning is insufficient and partitions them into sections of constant $C$-space dimensionality and robot modes. For each problem area, cross-sectional roadmaps are constructed to derive guide paths, which are then refined using standard motion planners. Motion-planning failures are quantified and fed back into the scheduling stage to iteratively improve candidate schedules.

The results show that the proposed approach reduces assembly duration compared with existing methods by enabling coordinated multi-robot motions. Its applicability is broad, provided that core operations have known durations and time-dependent collision geometries.

A main limitation is the assumption of fixed core-operation motions, which restricts the optimization potential. Although six scheduling methods were investigated, no single method consistently outperformed the others. In addition, computation times, while suitable for adaptive assembly scenarios, remain too high for higher-level iterative optimization involving changes in structural configuration or step assignment, and they scale nonlinearly with the number of robots. The resulting motion plans are geometric paths in the corresponding $X$-space. While velocity and acceleration constraints are partially incorporated, the timed geometric paths are not immediately executable on physical systems. 

Future work should focus on integrating the proposed method into a unified optimization framework that jointly considers structural and procedural configuration.




\bibliographystyle{IEEEtran}
\bibliography{bib/IEEEabrv,bib/IEEEexample}

\end{document}